\pdfoutput=1

\documentclass[11pt]{article}

\usepackage[final]{acl}

\usepackage{times}
\usepackage{latexsym}

\usepackage[T1]{fontenc}

\usepackage[utf8]{inputenc}

\usepackage{microtype}

\usepackage{inconsolata}

\usepackage{graphicx}
\usepackage{CJKutf8}
\usepackage{multirow}
\usepackage{array}
\usepackage{tabularx}
\usepackage{makecell}
\usepackage{amsmath}
\usepackage{longtable}
\usepackage{amssymb}
\usepackage{tablefootnote}
\usepackage{subcaption}
\usepackage{diagbox}
\usepackage{graphicx}
\usepackage{algorithm}
\usepackage[noend]{algpseudocode}
\usepackage{booktabs}
\usepackage{xcolor}
\usepackage{color, soul} 
\usepackage{colortbl}
\usepackage[raster,skins]{tcolorbox} 
\usepackage{xltabular}
\definecolor{orange}{rgb}{1,0.5,0}
\DeclareSymbolFont{extraup}{U}{zavm}{m}{n}
\DeclareMathSymbol{\varheart}{\mathalpha}{extraup}{86}
\DeclareMathSymbol{\vardiamond}{\mathalpha}{extraup}{87}
\usepackage{enumitem}

\usepackage{listings}

\newcolumntype{L}[1]{>{\raggedright\arraybackslash}p{#1}}
\newcolumntype{C}[1]{>{\centering\arraybackslash}p{#1}}
\newcolumntype{R}[1]{>{\raggedleft\arraybackslash}p{#1}}

\setlist[itemize]{topsep=3pt, partopsep=0pt, parsep=0pt, itemsep=3pt}

\title{Ziya2: Data-centric Learning is All LLMs Need}

\author{
    Ruyi Gan$^{\varheart\spadesuit}$ \qquad
    Renliang Sun$^{\varheart}$ \qquad
    Ziwei Wu$^{\varheart}$ \qquad
    Junyu Lu$^{\varheart}$ \qquad
    Xiaojun Wu$^{\varheart}$ \qquad
    \\
\hspace{8mm}\textbf{
    Dixiang Zhang$^{\varheart}$ \qquad
    Junqing He$^{\varheart}$ \qquad
    Yuanhe Tian$^{\spadesuit}$ \qquad
    Ping Yang$^{\varheart*}$ \qquad
    Qi Yang$^{\varheart*}$ \qquad
    }
    \\
\hspace{8mm}\textbf{
    Kunhao Pan$^{\varheart}$ \qquad
    Hao Wang$^{\varheart}$ \qquad
    Jiaxing Zhang$^{\varheart}$ \qquad
    Yan Song$^{\spadesuit}$ \qquad
    }
    \\
    \\
\hspace{2mm}
    $^\varheart$International Digital Economy Academy \quad
    $^\spadesuit$University of Science and Technology of China \quad
    \\
    \hspace{2mm}{\tt \{zhangjiaxing, wuziwei, sunrenliang\}@idea.edu.cn } \\
    {\tt ganruyi@mail.ustc.edu.cn, yhtian@uw.edu, clksong@gmail.com} \hspace{4mm}
}

\begin{document}
\maketitle
\begin{abstract}

Various large language models (LLMs) have been proposed in recent years, including closed- and open-source ones, continually setting new records on multiple benchmarks.
However, the development of LLMs still faces several issues,
such as high cost of training models from scratch, and continual pre-training leading to catastrophic forgetting, etc.
Although many such issues are addressed along the line of research on LLMs, an important yet practical limitation is that many studies overly pursue enlarging model sizes without comprehensively analyzing and optimizing the use of pre-training data in their learning process,
as well as appropriate organization and leveraging of such data in training LLMs under cost-effective settings.
In this work, we propose Ziya2, a model with 13 billion parameters adopting LLaMA2 as the foundation model, and further pre-trained on 700 billion tokens,
where we focus on pre-training techniques and use data-centric optimization to enhance the learning process of Ziya2 on different stages. We define three data attributes and firstly establish data-centric scaling laws to illustrate how different data impacts LLMs.
Experiments show that Ziya2 significantly outperforms other models in multiple benchmarks especially with promising results compared to representative open-source ones.\footnote{Ziya2 (Base) is released at \url{https://huggingface.co/IDEA-CCNL/Ziya2-13B-Base} and \url{https://modelscope.cn/models/Fengshenbang/Ziya2-13B-Base/summary}}






\end{abstract}

\section{Introduction}


LLMs have achieved great success in the field of artificial intelligence (AI), especially natural language processing (NLP), over the past few years.
Generally, LLMs are pre-trained on large amounts of text and show promising performance in a variety of NLP tasks without requiring intensive task-specific tuning on huge amounts of labeled data \cite{DBLP:conf/naacl/DevlinCLT19/bert,raffel2020exploring,joshi-etal-2020-spanbert,qin-etal-2021-relation,wang2022towards,lu2022unified,tian-etal-2023-end,ping-etal-2023-uniex,huang-etal-2023-mvp}.
Among all LLMs, representative ones include GPT-3 \cite{brown2020language} and its successors ChatGPT \cite{chatgpt}, GPT-4 \cite{openai2023gpt4}, and PaLM-2 \cite{anil2023palm}, demonstrate strong adaptability and applicability.
However, owing to the fact that the aforementioned LLMs are developed in rather restricted environments,
the lack of access to developers' source code and parameters becomes a barrier for many following researchers and developers in continuing LLM research based on existing well-performed models.
This paradox leads to the phenomenon that many researchers turn to train open-source alternatives, such as LLaMA2 \cite{touvron2023llama} and Falcon \cite{penedo2023refinedweb}, etc.,
since the open-source LLMs provide a foundation for further learning and improvement on the shoulders of successful predecessors, which promotes transparency and accountability in following AI studies.

\begin{figure*}[t!]
    \centering
    \includegraphics[width=0.95\linewidth, trim=9 5 0 -10]{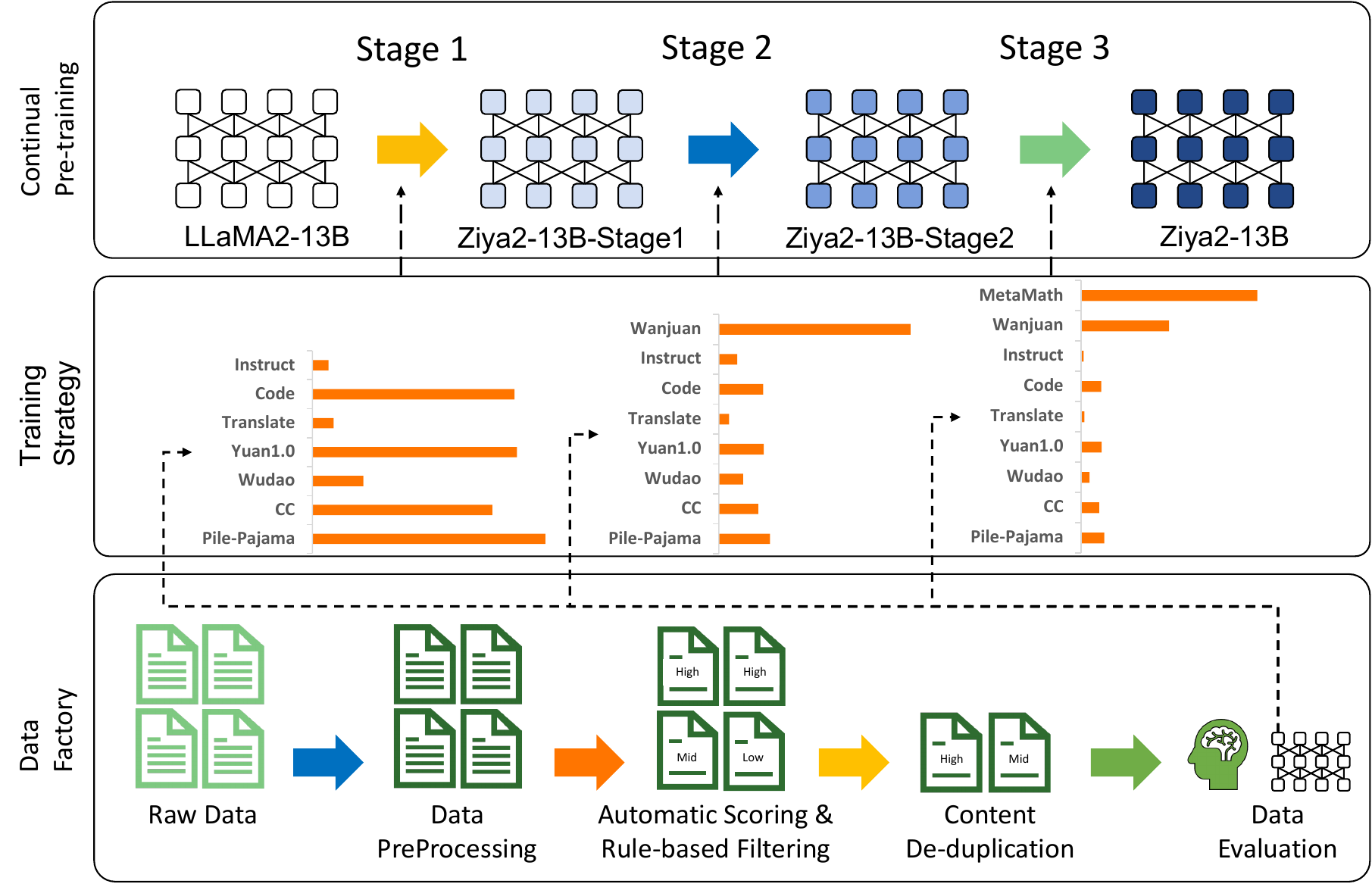}
    \caption{
    The overall data-centric process for training Ziya2, where the pipeline to obtain high-quality data, the training strategies, and the three-stage training process are presented in the bottom, middle, and top parts of the figure, respectively.
    The training strategies for different stages are illustrated by comparing the data distributions.
    }
    \label{fig:pipeline}
    \vspace{-0.3cm}
\end{figure*}

However, in addition to the various benefits that open-source LLMs bring, the development of LLMs is currently facing the following three major challenges.
The first one is that pre-training models from scratch necessitate long-term training on many GPUs, rendering the training of LLMs an extremely high-cost process. Continual pre-training presents a relatively cost-effective solution, yet it may be accompanied by issues such as catastrophic forgetting, which could be attributed to the disparities in data distribution.
The second one is that open-source LLMs often do not come with open-source data. The effectiveness of data processing methods directly impacts the performance of LLMs. Currently, there is no standardized methodology or criteria for cleaning pre-training data.
The third one is that many studies on LLMs prioritize expanding the capacity of model parameters and pre-training data to enhance model performance, often overlooking the impact of the quality of pre-training data on model performance.
Specifically, to our best knowledge, there is no study investigating which attributes of pre-training data exert the most significant influence on the LLMs. Under constraints of computational budget and a certain number of parameters, it is also worth exploring which type of data should be prioritized.


In this work, we focus on the continual pre-training strategies and understanding the intricate relationship between data and model performance. First, we propose a new data processing pipeline consisting of five steps to derive high-quality pre-training data from a vast corpus. We also involve human evaluation to ensure the quality of the processed data.
Then, we use LLaMA2-13B as the base model and propose a three-stage continual pre-training strategy. The first training stage utilizes lots of unsupervised data in English and Chinese. The second stage uses a relatively small amount of data but contains many supervised datasets. The third stage employs little augmented data that focuses on improving math abilities. Finally, we obtain Ziya2, and the whole data-centric training process is shown in Figure \ref{fig:pipeline}.

The evaluation of Ziya2 is performed on several representative benchmarks, and the results show that Ziya2 has a significant improvement over LLaMA2. Particularly,
Ziya2 improves LLaMA2 by \textbf{10\% on MMLU}, \textbf{61\% on CMMLU}, and \textbf{68\% on C-Eval}, respectively.
Especially in terms of mathematical benchmarks, Ziya2 improves LLaMA2 by \textbf{138\% on GSM8K} and \textbf{120\% on MATH},
and for programming benchmarks, Ziya2 improves it by \textbf{89\% on HumanEval}. Compared to other open-source models of the same size, Ziya2 also demonstrates superior performance. \textbf{It is shown that Ziya2 achieves outstanding performance on multidisciplinary datasets}\footnote{Detailed results are reported in Table \ref{tab:overall}.}, where, especially, it surpasses all open-source models used for comparison in mathematics and programming skills. Notably, Ziya2's performance on Chinese tasks also surpasses the GPT-3.5-turbo\footnote{We use the model under its ``\textit{gpt-3.5-turbo-0613}'' version.}. Our analysis of the intermediate checkpoints indicates that the impact of data from different stages on the model capabilities varies. Consequently, we conduct additional experiments and establish new data-centric scaling laws. We draw two conclusions suggesting that improving the `Coherence' and `Readability' of data is more effective in improving LLMs' capabilities than improving `Similarity' to downstream tasks.

In summary, our contributions include:
(1) We propose a new data processing pipeline and contribute over 700 billion tokens of high-quality data.
(2) We propose an effective continual pre-training strategy and obtain the Ziya2 model. We also perform a detailed analysis of the impact of the three stages' data on intermediate checkpoints.
(3) We define three data attributes and \textbf{firstly} establish the data-centric scaling laws for future research.

\section{Data Factory}
\label{sec:data_factory}

\begin{figure}[t]
    \centering
    \includegraphics[width=1\linewidth, trim=0 50 0 0]{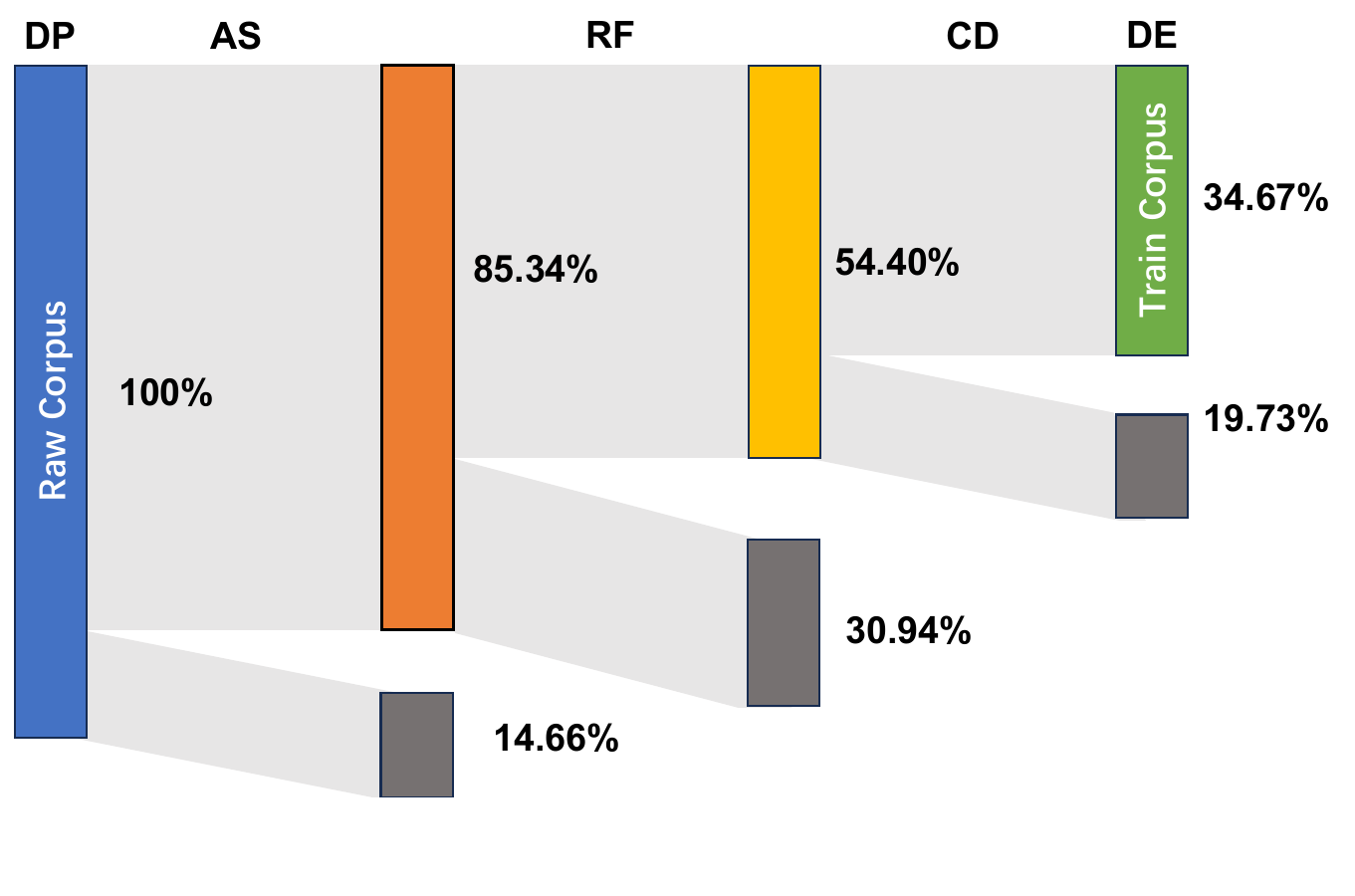}
    \caption{
    An overview of the proposed data processing pipeline, including five components: data preprocessing (DP), automatic scoring (AS), rule-based filtering (RF), content deduplication (CD), and data evaluation (DE).
    The gray and colored blocks indicate the proportion of data that is filtered out and retained relative to the original dataset, respectively.}
    \label{fig:enter-label}
    \vspace{-0.3cm}
\end{figure}

Data serves as the fundamental cornerstone for LLMs, with the scale and quality of the data determining the effectiveness of these models. In this work, we introduce a new data processing pipeline, as depicted in Figure \ref{fig:enter-label}. Ultimately, we have obtained high-quality data exceeding 700 billion tokens in scale.

\subsection{Data Processing Pipeline}
\label{sec:data_processing_pipeline}

This pipeline encompasses a variety of functions, including data preprocessing (DP), automatic scoring (AS), rule-based filtering (RF), content deduplication (CD), and data evaluation (DE):

\paragraph{Data Preprocessing (DP)}

Initially, we conduct a language detection on the collected corpus, selecting only English and Chinese data. Subsequently, we standardize the encoding of the corpus and convert all Traditional Chinese text into Simplified Chinese text. Following this, we eliminate meaningless tokens, such as invisible control characters, special symbols, emoticons, and so on.




\begin{table*}[ht]
\centering
\begin{tabular}{l|ccl}
\toprule
\textbf{Metric}          & \textbf{Method}                & \textbf{Level}          & \multicolumn{1}{c}{\textbf{Comments}}          \\ 
\midrule
Privacy Word Rate       & Re Matching           & 1              & \small{Includes email addresses, phone numbers and ect.}  \\
Toxic Word Rate         & Re Matching           & 1              & \small{Checking political, explicit, violent, or similar content.}  \\
Adv Word Rate           & Re Matching           & 1              & \small{Checking for the presence of advertising keywords.}  \\
Webpage Funcword Rate   & Re Matching           & 1              & \small{Detecting the presence of HTML tags.}  \\
Perplexity              & LM                    & 2              & \small{Whether the sentences in the article are coherent.}\\
Informativeness         & LM                    & 2              & \small{Assessing the information content within the text.}  \\
Readability             & Counts                & 2              & \small{Assessing text readability}  \\
Language                & LM                    & 3              & \small{Analyzing the distribution of languages.}  \\
Doc Length              & Counts                & 3              & \small{Evaluating the distribution of text lengths.}  \\
Topic                   & Statistics            & 3              & \small{Detecting the topic distribution of the text.}  \\
\bottomrule

\addlinespace[0.1cm]
\multicolumn{4}{c}{Automatic Evaluation} \\
\addlinespace[0.1cm]

\toprule                        
\textbf{Metric}                  & \textbf{Method}          & \textbf{Level}          & \multicolumn{1}{c}{\textbf{Comments}}          \\
                        \midrule
Coherence              & Human Check           & 1              & \small{Assessing consistency and usefulness.} \\
Readability             & Human Check           & 1              & \small{Assessing grammatical and formatting correctness.}  \\
Toxic        & Human Check           & 1              & \small{Checking political, explicit, violent, or similar content.} \\
\bottomrule
\addlinespace[0.1cm]
\multicolumn{4}{c}{Human Evaluation}
\end{tabular}%
\vspace{-0.2cm}
\caption{The metrics we utilize for evaluating data quality. The \textbf{Method} denotes our detection method, \textbf{Re Matching} signifies the method involving regular expressions for counting. \textbf{LM} represents the utilization of a language model for predicting relevant metrics. \textbf{Counts} indicates the use of statistical methods for directly countable metrics. \textbf{Human Check} indicates manual spot-checking conducted by humans. \textbf{Level} indicates the degree of stringency we apply to the respective metrics. We consider it compliant if it does not exceed one in a thousand.}
\label{tab:evaluate metric}
\vspace{-0.3cm}
\end{table*}

\paragraph{Automatic Scoring (AS)}

For the preprocessed data, we perform an automatic quality control (filtering) by language model scoring. We employ KenLM \cite{KenLM} to train two models seperately from the Chinese Wikipedia and the English Wikipedia and use them to evaluate the perplexity (PPL) of the data. We rank the PPL score from low to high. The top 30\% of data in terms of PPL ranking is considered high-quality data, and data with PPL ranking between 30\% and 60\% is deemed medium-quality data. We only keep high-quality data and medium-quality data.


\begin{table*}[t]
\centering
\begin{tabular}{l|l|cccc}
\toprule
             \textbf{Pre-training Stage} & \textbf{Dataset} & \textbf{Language}           & \textbf{Size}          & \textbf{Doc \#}          & \textbf{Token \#}                  \\ 
              \midrule
\multirow{7}{*}{Stage 1} & Pile-Pajama   & en                 & 400GB                   & 47M          & 110B               \\
& CC            & en                 & 384GB                    & 52M          & 109B                \\
& Wudao    & zh                 & 156GB                    & 51M         & 48B       \\
& Yuan1.0       & zh                 & 590GB                     & 260M         & 193B                 \\ 
& Translate     & multi              & ~~~~3GB                      & 12M          & 1.5B                 \\
& Code          & multi              & 480GB                    & 124M         & 191B                  \\
& Instruct      & multi                 & ~1.6GB                    & 0.9M             & 0.8B                \\
              \midrule
 Stage 2 &  Wanjuan  & zh                 & ~~76GB                     & 16M          & 29B                 \\
              \midrule
Stage 3 & MetaMath & en                 & ~0.3GB                    & 0.4M             & 0.1B                    \\
              \bottomrule
\end{tabular}%
\vspace{-0.2cm}
\caption{
The statistics of the obtained high-quality pre-training datasets. ``en'', ``zh'', and ``multi'' indicate that the main language of the dataset is English, Chinese, and multilingual, respectively.
}
\label{tab:data distrabute}
\vspace{-0.4cm}
\end{table*}



\paragraph{Rule-based Filtering (RF)}

Data mined from the internet may contain a substantial amount of explicit, terrorist, and discriminatory text, posing ethical challenges for LLMs \cite{zhuo2023exploring, touvron2023llama}.
Therefore, we design more than 30 filtering rules at three levels: document, paragraph, and sentence. At the document level, rules are designed around content length and format. At the paragraph and sentence levels, the focus of the rules shifts towards the toxicity. Notably, in the initial stages of rule design, we also randomly sample some original text for human evaluation.
We then update the rules based on the human feedback to ensure the effectiveness of the data filtering process.

%


\vspace{-0.1cm}

\paragraph{Content Deduplication (CD)}

Duplicate text within the pre-training data can detrimentally impact the performance of LLMs \cite{lee2021deduplicating,D4,penedo2023refinedweb}.
Thus, we use Bloomfilter \cite{bloomfilter} and Simhash \cite{simhash} to deduplicate the text. We also use caching and bucketing techniques to optimize this process. The detailed deduplication steps are as follows:
First, we use Bloomfilter to deduplicate text based on URLs, which significantly reduces the computational load required for subsequent content deduplication. Second, we use Bloomfilter to perform precise deduplication on these web pages. Finally, we employ SimHash for the fuzzy deduplication of textual content. Although this step may exclude some high-quality data, the evaluation of the sampled deduplicated data indicates that this loss is acceptable in the pursuit of greater training efficiency.

%
%

\vspace{-0.1cm}
\paragraph{Data Evaluation (DE)}

The last step of the pipeline is to evaluate the processed data. We evaluate the data quality using both automatic and human evaluations with specific metrics in Table \ref{tab:evaluate metric}. We also give the criteria for human annotations in Figure \ref{fig:annotations_criteria}. For automatic evaluation, we randomly select 1\% of the processed data to conduct evaluation. For human evaluation, we randomly sample 1,000 instances from the processed data and hire workers to evaluate them. Based on the automatic and human evaluations, we determine whether an example meets our data quality criteria. Then, we compute the rate of the unqualified examples over all evaluated instances.
If the rate is lower than a threshold, we consider these data to meet our quality criteria and include them in the training datasets.
If the rate is higher than the threshold, we consider the data does not meet our criteria. We will be more rigorous with processes like automatic scoring and rule-based filtering, and then conduct evaluations until the processed data meets the criteria.


\subsection{High-quality Pre-training Data}

We use the proposed data processing pipeline to clean multiple pre-training datasets: Pile-Pajama, CC, Wudao \cite{yuan2021wudaocorpora}, Yuan1.0 \cite{wu2021yuan}, Wanjuan \cite{he2023wanjuan}, MetaMath \cite{yu2023metamath} and pre-training data collected by ourselves, including code and book content. The details are as follows:

\begin{itemize}[leftmargin=0.3cm]

\item \textbf{Pile-Pajama} is a deduplicated fusion of Pile \cite{gao2020pile} and Redpajama \cite{redpajama} datasets after removing data from Common Crawl. 
\item \textbf{CC} is the Common Crawl \footnote{\url{https://commoncrawl.org/}} data from Pile and Redpajama. 
\item \textbf{Wudao} \cite{yuan2021wudaocorpora} is a super-large Chinese corpus containing about 3TB of training data.
\item \textbf{Yuan1.0} \cite{wu2021yuan} is an open-source dataset provided by Inspur Technology with more than 1TB of training data.
\item \textbf{Translate} is a multilingual translation dataset collected by ourselves. 
\item \textbf{Code} is a code dataset we collect from GitHub, which includes multiple programming languages such as C, C++, and Python. We add the program language type before the code and change it to a format that the Markdown syntax is able to recognize.
\item \textbf{Instruct} is a single-turn instructive dialogue dataset collected by ourselves.
\item \textbf{Wanjuan} \cite{he2023wanjuan} is a multilingual dataset from many different sources containing more than 1TB of training data. 
\item \textbf{MetaMath} \cite{yu2023metamath} is a dataset that achieves data augmentation by rewriting mathematical problems from multiple perspectives.
\end{itemize}

We also add some Chinese and English prompts for Instruct, Wanjuan, and MetaMath datasets, such as ``\textit{Q-A}'', ``\textit{question-answer}'', ``\textit{problem-solution}'', etc.


Table \ref{tab:data distrabute} shows the statistics of the high-quality pre-training dataset we constructed. We use different pre-training data in the three continual pre-training stages of LLMs. The data in the first stage contains solely unsupervised data. The second stage incorporates a substantial amount of supervised data. The data in the third stage is an augmentation of data specifically for mathematical tasks. We will elaborate on how to use them in the following section. Some pre-training data examples are presented in Appendix \ref{sec:pretraining_data}.

%



\section{Training Strategy}

In this section, we briefly introduce the proposed training strategy. We continue pre-training LLaMA2 \cite{touvron2023llama} on the constructed high-quality dataset using a novel three-stage training strategy.

\begin{figure}[ht!]
    \centering
    \includegraphics[width=1\linewidth, trim=0 10 0 0]
{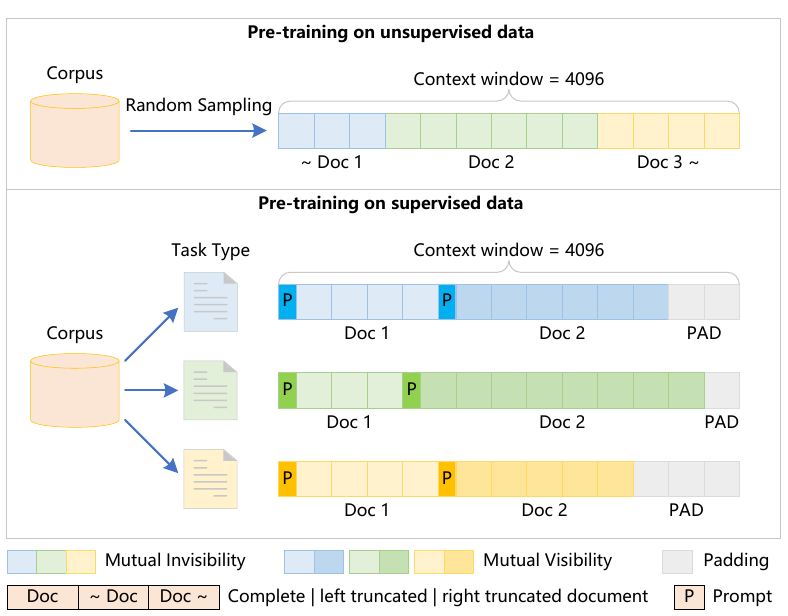}
    \caption{
    An illustration of how to process unsupervised and supervised data in the three training stages.
    }
    \label{fig:ziya2_data_construction}
    \vspace{-0.3cm}
\end{figure}

LLaMA2 performs excellently on general English benchmarks such as MMLU \cite{mmlu} but performs poorly in solving math and code programming problems. In addition, the fact that the pre-training data of LLaMA2 only contains little Chinese corpus also causes it to perform poorly on the Chinese benchmarks.
We want to enhance the Chinese, mathematical, and code programming capabilities of LLaMA2 while maintaining or improving its performance in English. To avoid the issue of catastrophic forgetting, we propose a three-stage training strategy.

In the first training stage, the training data consists of 650 billion tokens and is unsupervised. We randomly splice different documents together, using the special token `<eos>' to denote the end of a document. Text exceeding a length of 4,096 is truncated, as depicted in the upper half of Figure \ref{fig:ziya2_data_construction}.



In the second stage, we train the model using supervised task data with English and Chinese prompts. 
The supervised tasks include classification, reasoning, knowledge of single-choice \& multiple-choice, and question-answering. We design specific prompts for each task and concatenate these prompts with task documents.
We still use the next token prediction method for continual pre-training. However, in contrast to the first stage, we concatenate documents of the same task type. Furthermore, we ensure that the length of the concatenated text is close to but does not exceed 4,096, with the remaining part filled with the `<pad>' token. This process is shown in the lower half of Figure \ref{fig:ziya2_data_construction}. To maintain the English capability of the model, we randomly sample data of the same size from the first stage.


In the third stage, we incorporate the supervised dataset MetaMath \cite{yu2023metamath} to improve the mathematical reasoning abilities of the model, using the same composition approach as in the second stage. We also randomly sample 0.1B tokens from the first and the second stage data to mix with MetaMath, respectively. Finally, we obtain the Ziya2 model.


\section{Improvements to the Structure of LLaMA2}
\label{sec:appendix_improvement}
The Ziya2 architecture is built on top of LLaMA2, and aims to enhance the quality of input data processing, token and hidden representations.
The details of these improvements are described in the following sections.

\subsection{Tokenizer}
\label{sec:tokenizer}
To improve the preservation of semantic meaning in text and provide better support for Chinese, we have adopted a BPE (Byte-Pair Encoding) tokenizer \cite{bpe}.
The vocabulary of the tokenizer includes over 600 Chinese tokens originally used in LLaMA2, as well as an additional 7,400 commonly used simplified and traditional Chinese tokens. The reason for adding extra Chinese tokens is that the original LLaMA2 vocabulary with BPE tokenizer is not efficient for Chinese, primarily due to how computers handle character encoding. For instance, in most cases, one UTF-8 Chinese character is encoded into 2-4 tokens using BPE encoding. After adding Chinese tokens to the vocabulary, testing on a 10GB Chinese corpus shows an efficiency improvement of approximately 34\% in Chinese encoding and decoding compared to the original LLaMA2 tokenizer.

\subsection{Positional Embedding}
During continual pre-training, we observe a divergence in the distribution of text lengths between our continued pre-training dataset and the LLaMA2 pre-training dataset. This necessitates an adaptation of the position embedding to different data distributions.
Meanwhile, considering downstream tasks involving lengthy texts, the scalability of position embedding is of significant importance. 
LLaMA2 employs rotary position encoding \cite{su2021roformer}, which, through the mechanism of absolute position encoding, accomplishes relative position encoding. 
To avoid the overflow issues associated with mixed precision, we implement rotary position encoding using FP32 precision, thereby accommodating the variation in data length distribution in continual pre-training.

\begin{figure*}[t]
    \centering
    \includegraphics[width=1\linewidth, trim=0 10 0 0]{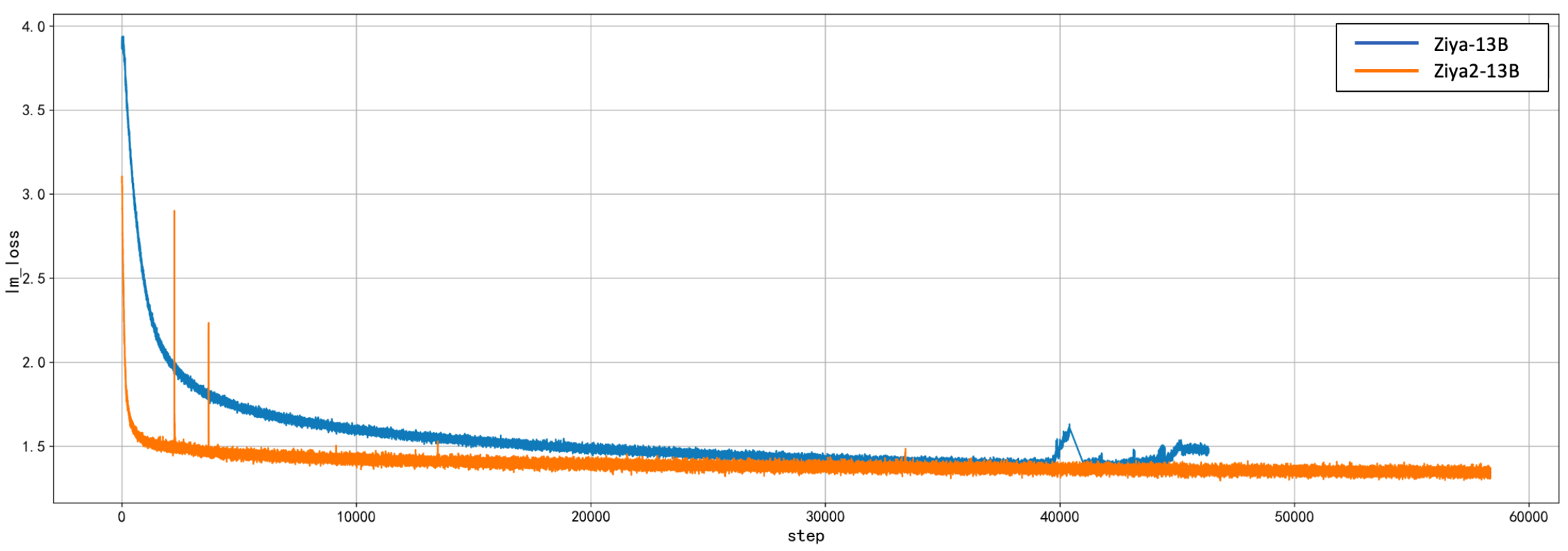}
    \caption{The pre-training loss of Ziya2-13B and Ziya-13B with respect to the number of training steps.}
    \label{fig:ziya2_loss}
    \vspace{-0.3cm}
\end{figure*}

\subsection{Layer Normalization and Attention}
We find that the direct implementation of mixed precision training in layer normalization and attention leads to precision overflow, which results in training instability. In order to maintain the efficiency and stability of model training, we propose structural improvements to layer normalization and attention in LLaMA2. For layer normalization, we utilize an APEX \footnote{\url{https://github.com/NVIDIA/apex}} RMSNorm \cite{RMSNorm} implementation that operates under FP32 precision training. For attention, we employ a fused operator to replace the original scaling, mask, and softmax operators within the attention module, thereby expediting the computation of attention. To prevent overflow during mixed precision training in the softmax operator, it is trained using FP32 precision. These structural improvements enhance the training efficiency of the model and better adapt the training process to the instability brought about by changes in data distribution.





%



\section{Training Details}
\label{sec:appendix_training_details}

\subsection{Initialization}
As mentioned in Section \ref{sec:tokenizer}, we expand the vocabulary by 7,400 Chinese tokens. These tokens can be represented by 2-4 tokens of the vocabulary of LLaMA2. Thus, we calculate the weighted average embeddings of the corresponding tokens of the vocabulary of LLaMA2 as the initial embeddings of the new Chinese tokens.

\subsection{Optimizer}
The AdamW optimizer \cite{adamw} with hyperparameters $\beta_1 = 0.9$ and $\beta_2 = 0.95$ is used to train the model. Our findings indicate that incorporating additional Chinese and code data in our pre-training dataset compared to LLaMA2 results in a disparity in the overall data distribution. To address this issue, we propose a more extended warmup for continual pre-training. Specifically, we adopt a warmup ratio of 1\% instead of the 0.4\% warmup ratio utilized in LLaMA2. This is followed by a cosine learning rate decay schedule, which reaches a final learning rate of $1e^{-5}$. We also implement a weight decay of 0.1 and gradient clipping set at 1.0 to enhance the training efficiency of the model.

\subsection{Training Frameworks}
We have employed Megatron \cite{shoeybi2019megatron} and DeepSpeed \cite{rasley2020deepspeed} as foundational training frameworks. Megatron enables distributed training of large-scale models through data parallelism, tensor parallelism, and pipeline parallelism. We also utilize the ZeRO technology \cite{rajbhandari2020zero} from DeepSpeed to optimize memory savings. We trained the model with 80 NVIDIA A100-80G GPUs, setting the value of model parallelism to 1 and pipeline parallelism to 2. We have implemented multiple advanced techniques such as flash-attention \cite{dao2022flashattention} and fused-softmax to enhance training efficiency. As a result, each GPU achieves an industry-leading efficiency of 163.0 TFLOPS per second.

\begin{figure}[ht!]
    \centering
    \includegraphics[width=1\linewidth, trim= 0 10 0 0]{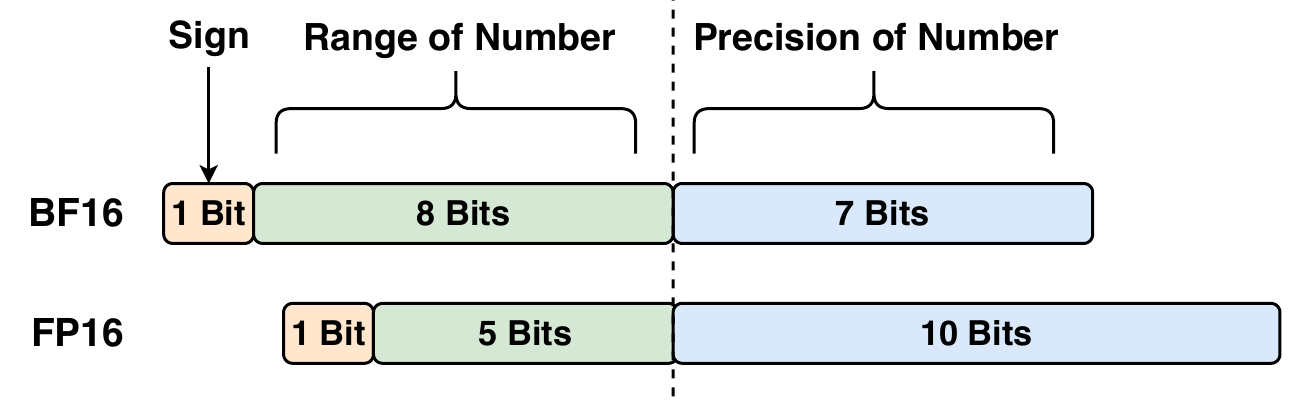}
    \caption{An illustration of the representations of floating numbers by BF16 and FP16.}
    \label{fig:bf16andFP16}
    \vspace{-0.3cm}
\end{figure}

\subsection{Training Stability}

We notice that when using the FP16 mixed precision \cite{micikevicius2017mixed} to train the Ziya-13B model, a loss spike occurs in the later stages of training. We find that the limited numerical range of FP16 leads to overflow problems, especially in operations such as softmax. Compared to FP16, BF16 has a larger representation range, as shown in Figure \ref{fig:bf16andFP16}. Hence, we opt for BF16 mixed-precision training for the continual pre-training of Ziya2-13B. We also fix some underlying bugs in DeepSpeed to address the loss spike problems. As a result, our model achieves convergence even when training continues for 700B tokens, as shown in Figure \ref{fig:ziya2_loss}.

\section{Results}

\begin{table*}[ht!]
\centering
\scalebox{0.96}{
\begin{tabular}{l|l|ccc|ccc}
\toprule
\textbf{Source} & \textbf{Model} & \textbf{MMLU}           & \textbf{CMMLU}          & \textbf{C-Eval}          & \textbf{GSM8K}          & \textbf{MATH}           & \textbf{HumanEval}      \\ \midrule
\multirow{2}{*}{Closed-source}
& GPT-4         & 83.93          & 70.33          & 68.40          & 89.99          & 40.20          & 69.51          \\
& GPT-3.5-turbo & 68.54          & 54.06          & 51.10          & 57.77          & 13.96          & 52.44          \\ \midrule
\multirow{7}{*}{Open-source} 
& ChatGLM2-6B   & 47.86          & 49.30              & 51.70          & 28.89          & ~~6.40           & ~~9.15           \\
& Falcon-7B     & 26.03          & 25.66          & 24.23          & ~~5.46           & ~~1.68           & -              \\
& Vicuna-13B    & 52.00          & 36.28          & 32.80          & 28.13          & ~~4.36           & 16.46          \\
& XVERSE-13B    & 55.21         & 58.44          & 53.70          & 18.20          & ~~2.18           & 15.85          \\
& WeMix-13B    & 59.70          & 42.60          & 42.70          & 45.20          & ~~6.90           & 24.40          \\
& Baichuan2-13B & 59.17          & \textbf{61.97} & 58.10          & 52.77          & 10.08          & 17.07          \\ 
& LLaMA2-13B    & 55.74          & 37.90          & 34.99          & 28.81          & ~~4.98           & 15.24          \\ \midrule
\multirow{2}{*}{Ours}
& Ziya-13B      & 43.88          & 31.09          & 28.97          & 17.97          & ~~3.08           & 14.02          \\
& Ziya2-13B    & \textbf{61.36} & 60.95          & \textbf{58.84} & \textbf{68.46} & \textbf{10.98} & \textbf{28.87} \\ \bottomrule
\end{tabular}%
}
\vspace{-0.2cm}
\caption{Comparison of Ziya2 with other closed-source and open-source LLMs on six benchmark datasets for LLM evaluation, where the \textbf{boldface} indicates the best-performing result over all open-source LLMs.}
\label{tab:overall}
\vspace{-0.3cm}
\end{table*}

\subsection{Evaluation Method}
\label{sec:evaluation}

Following previous research \cite{chatglm,touvron2023llama,baichuan2}, we evaluate LLMs on six representative benchmarks: MMLU \cite{mmlu}, CMMLU \cite{cmmlu}, C-Eval \cite{ceval}, GSM8K \cite{gsm8k}, MATH \cite{math}, and HumanEval \cite{humaneval}. MMLU tests the English general abilities of LLMs. CMMLU and C-Eval test the Chinese general abilities of LLMs. GSM8K and MATH test the math ability of LLMs. HumanEval tests the code programming ability of LLMs. We use the official evaluation scripts in OpenCompass \cite{contributors2023opencompass} to compute the accuracy and perform 5-shot evaluations. The details of the evaluation datasets are shown as follows:

\begin{itemize}[leftmargin=0.3cm]
    \item \textbf{MMLU} (Massive Multitask Language Understanding) \cite{mmlu} provides a comprehensive evaluation of models in both zero-shot and few-shot settings, spanning across a wide range of 57 subjects. The unique aspect of MMLU is that it tests the models' world knowledge and problem-solving capabilities.
    \item \textbf{CMMLU} (Chinese Massive Multitask Language Understanding) \cite{cmmlu} is an extensive evaluation suite tailored to measure the advanced knowledge and reasoning skills of LLMs in the context of the Chinese language and culture. It contains a broad spectrum of 67 topics, ranging from basic to advanced professional levels.
    \item \textbf{C-Eval} (Chinese Evaluation) \cite{ceval} is a thorough evaluation suite specifically designed for foundational models in Chinese. It is composed of 13,948 multiple-choice questions that cover a wide array of 52 different disciplines and span across four levels of difficulty.
    \item \textbf{GSM8K} (Grade School Math 8K) \cite{gsm8k} is a collection of 8,500 high-quality math word problems created by human writers. The objective of this dataset is to facilitate the task of question answering on fundamental problems that necessitate reasoning through multiple steps.
    \item \textbf{MATH} \cite{math} aggregates 12,500 intricate competition mathematics problems. A unique feature of this dataset is that each problem comes with a comprehensive step-by-step solution. These detailed solutions serve as a valuable resource for teaching models to generate derivation processes and explanations.
    \item \textbf{HumanEval} \cite{humaneval} is a meticulously constructed set of 164 programming challenges. This dataset serves as a benchmark for evaluating the ability of a system to generate functionally correct programs based on provided doc strings. The challenges encompass a range of topics, including language comprehension, algorithmic problems, and basic mathematics.
\end{itemize}

\subsection{Benchmark Results}

We choose multiple open-source models with similar sizes to Ziya2 as baselines: ChatGLM2 \cite{chatglm}, Falcon \cite{penedo2023refinedweb}, Vicuna \cite{vicuna}, Baichuan2 \cite{baichuan2}, XVERSE, WeMix, LLaMA2 \cite{touvron2023llama} and Ziya \cite{fengshenbang}. We also choose closed-source models GPT-3.5-turbo and GPT-4 \cite{openai2023gpt4} as references. The details about baselines are as follows:


\begin{itemize}[leftmargin=0.3cm]
    \item \textbf{ChatGLM2-6B} \cite{chatglm} is developed by Tsinghua University and Zhipu AI. The model is the second generation of the bilingual dialogue model ChatGLM-6B.
    \item \textbf{Falcon-7B} \cite{penedo2023refinedweb} is a causal decoder-only model developed by Technology Innovation Institute (TII). It's trained on 1,500 billion tokens of RefinedWeb.
    \item \textbf{Vicuna-13B} \cite{vicuna} is an open-source LLM developed by the Language Model Systems (LMSYS) organization and is fine-tuned from LLaMA on user-shared conversations between humans and ChatGPT.
    \item \textbf{Baichuan2-13B} \cite{baichuan2} is developed by Baichuan Intelligent Technology and is trained with 2.6 trillion tokens.
    \item \textbf{XVERSE-13B}\footnote{\url{https://github.com/xverse-ai/XVERSE-13B}.} is developed by XVERSE Technology and is trained with 1.4 trillion tokens.
    %
    \item \textbf{WeMix-13B}\footnote{\url{https://github.com/Alpha-VLLM/WeMix-LLM}.} is developed by Shanghai AI lab and is built on LLaMA2-Accessory.
    \item \textbf{LLaMA2-13B} \cite{touvron2023llama} is Meta's open-source large language model. It is designed for dialogue scenarios and is freely available for research and commercial use.
    \item \textbf{Ziya-13B} \cite{fengshenbang} is continue pre-trained on the LLaMA-13B model and performs well on many downstream tasks.
    \item \textbf{GPT-3.5-turbo} is a high-performance variant of GPT-3 and it is proficient in text completion.
    \item \textbf{GPT-4} \cite{openai2023gpt4} is the latest state-of-the-art LLM developed by OpenAI. It exhibits human-level performance on various professional and academic benchmark datasets.
\end{itemize}

The results of different LLMs on the six benchmarks are shown in Table \ref{tab:overall} with the following observations.

First, Ziya2 significantly outperforms LLaMA2 on all the benchmarks. Specifically, for general English tasks, Ziya2 outperforms LLaMA2 by 6 points on MMLU. For general Chinese tasks, Ziya2 surpasses LLaMA2 by 23 and 24 points on CMMLU and C-Eval, respectively. For specific downstream tasks, Ziya2 outperforms LLaMA2 by 40, 6, and 13 points on GSM8K, MATH, and HumanEval datasets, respectively.

\begin{figure*}[ht!]
    \centering
    \includegraphics[width=16cm, trim=0 15 0 0]{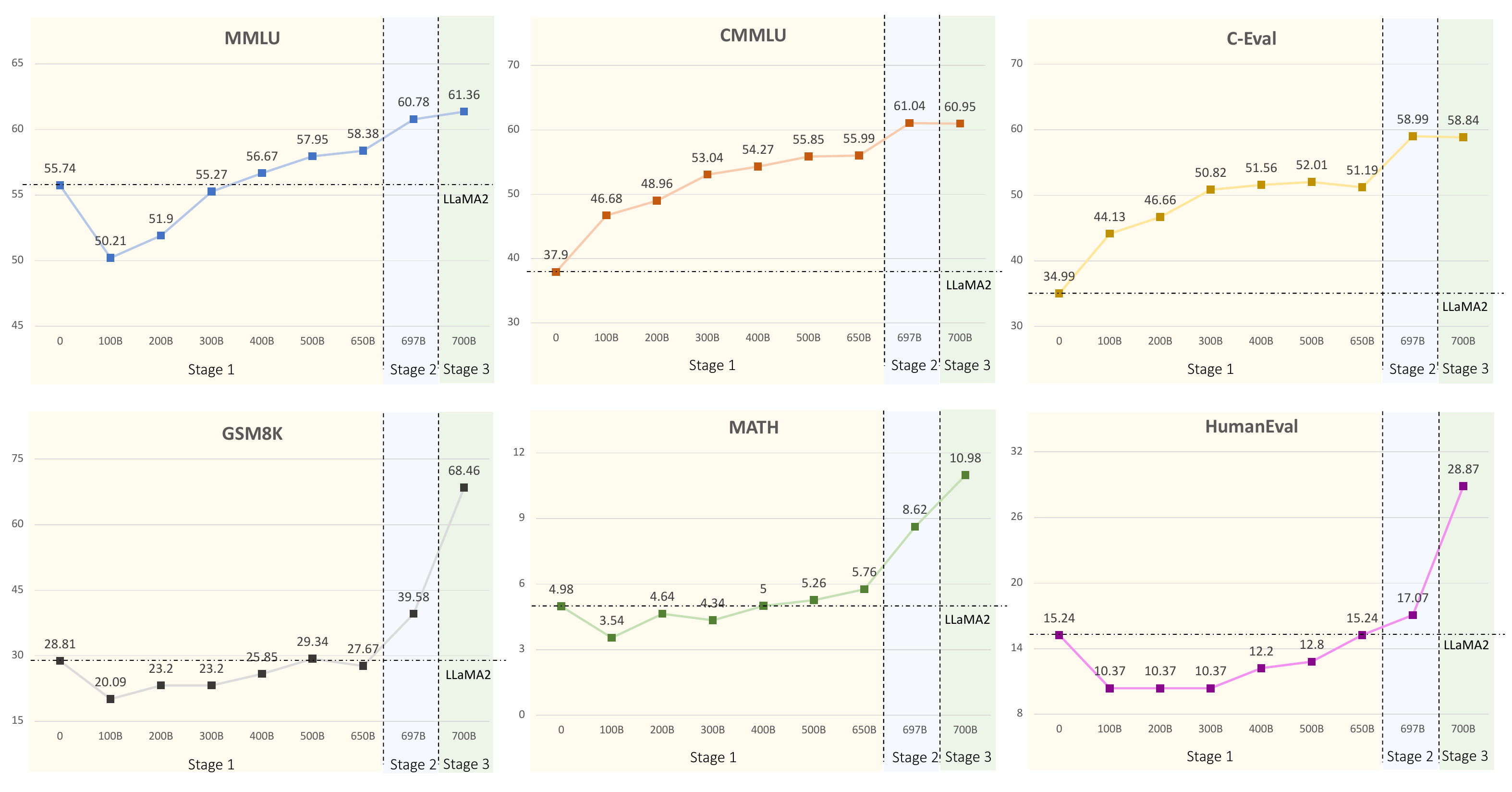}
    \caption{
    The performance of Ziya2 on the six benchmark datasets with respect to the three training stages.
    The training process in different stages is represented by the number of training tokens at a particular training step.
    The performance of LLaMA2 on the datasets is illustrated by dashed lines for reference.
    }
    \label{fig:checkpoints}
    \vspace{-0.3cm}
\end{figure*}

The results highlight the effectiveness of our continual pre-training strategy. It not only enhances LLaMA2's English capabilities and mitigates catastrophic forgetting but also significantly improves its performance in Chinese, mathematical, and code programming tasks.

Second, compared to open-source baselines of comparable size, Ziya2 outperforms almost all LLMs on all benchmarks.
Remarkably, Ziya2's Chinese proficiency is even superior than that of GPT-3.5-turbo, and Ziya2's mathematical ability is on par with GPT-3.5-turbo.
Although the Baichuan2-13B model performs comparably to Ziya2 on general tasks, Ziya2 significantly outperforms Baichuan2-13B in mathematical and code programming abilities. The results indicate that the Ziya2 model is at the forefront on the model scale of 13 billion parameters.

In Appendix \ref{sec:generation_cases}, we give examples generated by Ziya2 across three major domains: English, Chinese, and code programming.

\subsection{Data Efficiency}
\label{sec:data_efficiency}

In addition to the final results, we report the performance of Ziya2's intermediate checkpoints on six benchmarks in Figure \ref{fig:checkpoints}. This section presents the observations from the three training stages, emphasizing the different effects of data on LLMs' performance.
%

In the first training stage of continual pre-training, Ziya2's performance on MMLU deteriorates due to the inclusion of a large amount of Chinese corpus in the training data that is different from the setting of LLaMA2. However, as the number of training steps increases, Ziya2 learns from a broader view of more data, which enhances its capabilities in both Chinese and English text processing. In particular, new data significantly improves Ziya2's performance on CMMLU and C-Eval benchmarks for Chinese tasks that LLaMA2 is not optimized for. A modest enhancement is also synchronously observed in Ziya2's mathematical and code programming abilities.

In the second stage, Ziya2 exhibits a more substantial enhancement on six benchmarks relative to the first stage, with notable advancements on CMMLU, C-Eval, GSM8K, and MATH. 
The results may underscore the greater contribution of supervised data over unsupervised data to continual pre-training. Therefore, employing supervised data for pre-training is able to reduce the number of training steps and economize costs.

In the third stage, we observe that training with the data-augmented dataset significantly improves the performance of Ziya2 on GSM8K, MATH, and HumanEval.
On one hand, the experimental results prove that data augmentation specific to a particular dataset is able to significantly boost the model's performance on that dataset. On the other hand, such an ``effortless'' enhancement in model performance may not necessarily be beneficial, as the model might merely learn the format of the problems, rather than genuinely improving its mathematical capabilities. More details are given in the next section.

\section{Data-centric Scaling Laws}

%

We want to explore the effect of different data on the model's performance, keeping the size of the model constant. Previous research on the scaling laws simply represents data in terms of data volume \cite{kaplan2020scaling, hoffmann2022training, muennighoff2023scaling}. In this work, we employ two attributes, `Coherence' and `Readability' detailed in Table \ref{tab:evaluate metric}, along with the `Similarity' attribute to characterize the data. `Coherence' and `Readability' are also used to evaluate the quality of the pre-trained data in Section \ref{sec:data_processing_pipeline}. We discard the `Toxic' attribute due to the sparsity of its human evaluation results as showed in Table \ref{tab:attributes}. `Coherence' reflects the consistency and usefulness of the data content, `Readability' assesses the grammatical and formatting correctness of the data, and `Similarity' measures the extent of similarity between the pre-training data and the test set of downstream tasks. 

\begin{table}[ht!]
\centering
\begin{tabular}{l|ccc}
\toprule
\textbf{Attibute}       & \textbf{Stage1} & \textbf{Stage2}          & \textbf{Stage3}          \\ \midrule
Coherence     & 0.801 &	0.653 & 0.993 \\
Readability   & 0.917 &	0.975 &	0.916          \\
Toxic & 0.990 & 1.000 & 1.000 \\ \bottomrule
\end{tabular}
\caption{The values of `Coherence', 'Readability', and 'Toxic' of data from the three stages. We perform human evaluations, and the criteria are shown in Figure \ref{fig:annotations_criteria}.}
\label{tab:attributes}
\end{table}

\begin{table}[ht!]
\centering
\begin{tabular}{l|ccc}
\toprule
\textbf{Benchmark}       & \textbf{Stage1} & \textbf{Stage2}          & \textbf{Stage3}          \\ \midrule
MMLU      & 0.6714 & 0.6677          & \textbf{0.6827} \\
CMMLU     & 0.7057 & \textbf{0.7417}          & 0.6567          \\
C-EVAL    & 0.7024 & \textbf{0.7374}          & 0.6634          \\
GSM8K     & 0.6612 & 0.6570          & \textbf{0.7524} \\
MATH      & 0.6552 & 0.6505          & \textbf{0.7477} \\
HumanEval & 0.5689 & 0.5563          & \textbf{0.6211} \\ \bottomrule
\end{tabular}
\caption{The `Similarity' value between data from the three stages and the test sets of the six tasks. We use the OpenAI's text embeddings api and cosine similarity to calculate the `Similarity' value.}
\label{tab:similarity}
\end{table}

\citet{hoffmann2022training} propose Eq. \ref{eq:1}: to minimize the loss (L) under the computational resource constraints (C, measured in FLOPs) through an optimal allocation of the model size (N) and the data volume (D).

\begin{equation}
\begin{aligned}
{\rm \mathop{argmin}\limits_{N,D}}L(N,D) \quad {\rm s.t. \; FLOPs}(N,D)=C
\end{aligned}    
\label{eq:1}
\end{equation}

The loss (L) can be modeled as a parametric equation of the model size and the data volume, where \{A, $\alpha$, B, $\beta$, E\} are learnable parameters:

\begin{equation}
\begin{aligned}
L(N, D) =  \frac{A}{N^\alpha} + \frac{B}{D^\beta} + E
\end{aligned}   
\label{eq:2}
\end{equation}

\begin{figure*}[t]
    \centering
    \includegraphics[width=1\linewidth, trim=0 10 0 0]{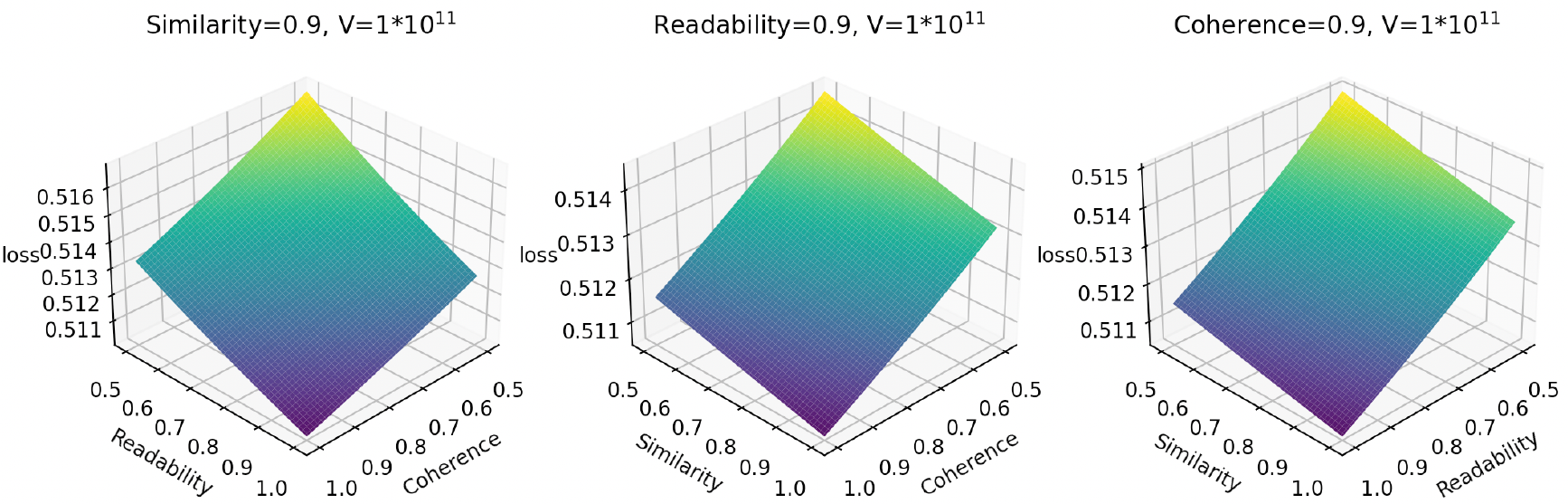}
    \caption{The different effects of three data attributes on the loss. The loss decreases at roughly the same rate as `Readability' and `Coherence' increase, but much faster than when `Similarity' increases.}
    \label{fig:three_figures_comparison}
    \vspace{-0.3cm}
\end{figure*}

Our goal is to introduce a modified version of Eq. \ref{eq:2} to account for the effects of different attributes of the data on the LLM's performance. In the modified equation, the loss (L) is modeled as a parametric equation of `Coherence (CH)', `Readability (RA)', `Similarity (SIM)', and the data volume ($V$), where \{$\alpha$, $A$, $B$, $C$, $E$, $F$\} are learnable parameters:

\begin{equation}
\begin{aligned}
L(D) = D^{-\alpha} + E
\end{aligned}
\label{eq:4}
\end{equation}

\begin{equation}
\begin{aligned}
D = (A * \rm{CH} \; + \; & B * \rm{RA} +  C * \rm{SIM}) \\ & * V + F
\end{aligned}
\label{eq:5}
\end{equation}

In contrast to Eq. \ref{eq:2} , we only conduct experiments on a fixed model size, so the term $\frac{A}{N^{\alpha}}$ can be omitted as a constant. $V$ represents the data volume to continual pre-training. When $V$ is 0, $L$ denotes the loss of the initial pre-trained model on the test set. $A$, $B$, $C$ are the coefficients of the data attributes. $E$ represents the loss for an ideal generative process on the data distribution. $F$ represents the effect of the pre-training data on the loss.


We perform human evaluations by taking 1,000 pieces of data from each of the three stages of continual pre-training data to obtain the values of `Coherence' and `Readability'. The results are shown in Table \ref{tab:attributes}. We then extract 100,000 pieces of data from each of the three stages and calculate the `Similarity' value of the data to the test set for each of the six downstream tasks, shown in Table \ref{tab:similarity}. 
Different from the experimental setup in Section \ref{sec:data_efficiency}, we continue pre-training only on data from the three stages, respectively. We have not mixed the data with other stages to avoid interference. Then, we calculate the test loss for the downstream tasks at various checkpoints with different training data volumes. Finally, we have obtained 84 samples, each characterized by three unique data attributes, data volumes, and test loss values.

To estimate \{$\alpha$, $A$, $B$, $C$, $E$, $F$\} in Eq. \ref{eq:4} and Eq. \ref{eq:5}, we minimize the Huber loss \cite{huber1964robust} between the observed log loss $L^{'}_{i}$ and the predicted log loss, where LSE is the log-sum-exp operator:


\begin{equation}
\begin{aligned}
& {\rm \mathop{min}\limits_{\alpha, A, B, C, e}} = \sum\limits_{{\rm Runs \; i}}{\rm Huber}_{\delta} \\ ( {\rm L} & {\rm SE} (1-\alpha logD_i, e) - logL^{'}_{i} ) 
\end{aligned}
\label{eq:6}
\end{equation}

In Eq. \ref{eq:6}, we set e = log(E) and $\delta = 10^{-3}$ for the Huber loss. We use the L-BFGS algorithm \cite{liu1989limited} to find the local minima of the objective above. 
For `Coherence' and `Readability', we extract 1,000 pieces of data from each of the stages for human evaluation. The results are then normalized to a range of [0,1]. For `Similarity', we randomly extract 100,000 pieces of data from each of the stages and compute their embeddings, respectively. We use text-embedding-ada-002 from OpenAI to calculate the embeddings. Subsequently, we calculate the embeddings for the test sets of six benchmarks in Section \ref{sec:evaluation}. The average cosine similarity between them is used as the value for `Similarity', with a range of [0,1].

The L-BFGS algorithm starts on a grid initialization given by: $\alpha \in \{0, 0.67, 1.33, 2\}$, $A \in \{0, 0.29, ... , 1.71, 2\}$, $B \in \{0, 0.29, ... , 1.71, 2\}$, $C \in \{0, 0.29, ... , 1.71, 2\}$, $E \in \{0, 0.67, 1.33, 2\}$, $F \in \{1, 1.67, 2.33, 3\}*10^{12}$.
Empirically, we find after fitting Eq. \ref{eq:4} and Eq. \ref{eq:5} that

\begin{equation}
\begin{aligned}
L(D) = D^{-0.06} + 0.31
\end{aligned}
\label{eq:7}
\end{equation}

\begin{equation}
\begin{aligned}
D = (1.66 * \rm{CH} \; + \; & 1.98 * \rm{RA} +  0.70 * \rm{SIM}) \\ & * V + 1.24*10^{10}
\end{aligned}
\label{eq:8}
\end{equation}

From Eq. \ref{eq:8}, it is evident that the coefficients of `Coherence' and `Readability' are much larger than that of `Similarity'. It suggests that to enhance the performance of LLMs, improving the semantic consistency and grammatical correctness of the pre-training data is more effective than increasing the similarity between the pre-training data and the downstream tasks. We visualize the effect of different data attributes on loss, as shown in Figure \ref{fig:three_figures_comparison}. Although previous work \cite{yu2023metamath, liu2023tinygsm} augment data for specific tasks such as mathematics, resulting in a substantial improvement in the model's performance on these tasks, our experimental results indicate that if the model's performance on general tasks is taken into account, enhancing the similarity between the data and the downstream tasks does not significantly improve the model's capabilities. Finally, we arrive at two conclusions: \textbf{(1) Improving the semantic and grammatical quality of pre-training data is more effective in enhancing model performance than data augmentation. 
(2) If data augmentation methods are used to obtain pre-training data, it is better to mix in high-quality generic data and report the model's performance on general tasks concurrently.
}

\section{Related Work}

In recent years, LLMs have made significant strides in artificial intelligence. GPT-3 \cite{brown2020language}, which uses only a decoder structure, represents an important milestone due to its enormous model size and impressive zero-shot and few-shot in-context learning performance.
InstructGPT \cite{instructgpt} further improves GPT-3 by supervised fine-tuning and reinforcement learning from human feedback (RLHF), where human-annotated data is used to train the LLM. Following InstructGPT, ChatGPT \cite{chatgpt} has gained significant attention from the public due to its remarkable text-processing capabilities. Later, GPT-4 \cite{openai2023gpt4} presents huge improvements over its predecessors and can process multimodal data, including both images and text.

In order to utilize and study LLMs, researchers are committed to developing open-source models since the parameters of the above LLMs are not publicly available.
LLaMA \cite{llama1} and LLaMA2 \cite{touvron2023llama} are two representative open-source LLMs used in many downstream tasks. LLaMA2 is mainly trained on English data and is trained on more tokens than LLaMA. In response to the subpar performance of open-source LLMs on Chinese tasks, many researchers propose new LLMs for Chinese text processing. For instance, Chinese-Vicuna \cite{chinesevicuna} is a low-cost Chinese dialogue model based on LLaMA. ChatGLM2 \cite{chatglm} supports dialogue in both Chinese and English, which is specifically optimized for Chinese dialogue tasks. Baichuan2 \cite{baichuan2}, QWEN \cite{qwen}, and Yi \cite{yi} have achieved commendable results on multi-disciplinary benchmarks in both English and Chinese.

\section{Conclusion}


In this paper, we propose Ziya2, an open-source LLM with 13 billion parameters for Chinese and English text processing.
Ziya2 is based on the open-source LLaMA2 model and is continually pre-trained through the proposed data-centric learning approach.
Specifically, we collect and clean open-source data from the internet, developing a comprehensive data processing system and accumulating terabytes of high-quality data, which is used to train Ziya2 through three stages. 
%
The performance of Ziya2 in Chinese and English downstream tasks not only surpasses LLaMA2 but also outperforms contemporaneous open-source LLMs of similar size, which demonstrates the effectiveness of the data-centric learning approach for LLM pre-training. Based on the experimental results, we explore the impact of different data attributes on LLMs and propose data-centric scaling laws.
%
%
In the future, we plan to continue training Ziya2, explore larger models with 70B parameters, and align Ziya2 to achieve better instructional compliance with the Ziya2-Chat model.
We aim to release specialized LLMs for various domains such as
writing, coding, and multimodality.
\bibliography{custom,ours}

\clearpage

\appendix

\clearpage

\begin{figure*}[t!]
    \centering
    \includegraphics[width=1\linewidth, trim=9 5 0 -10]{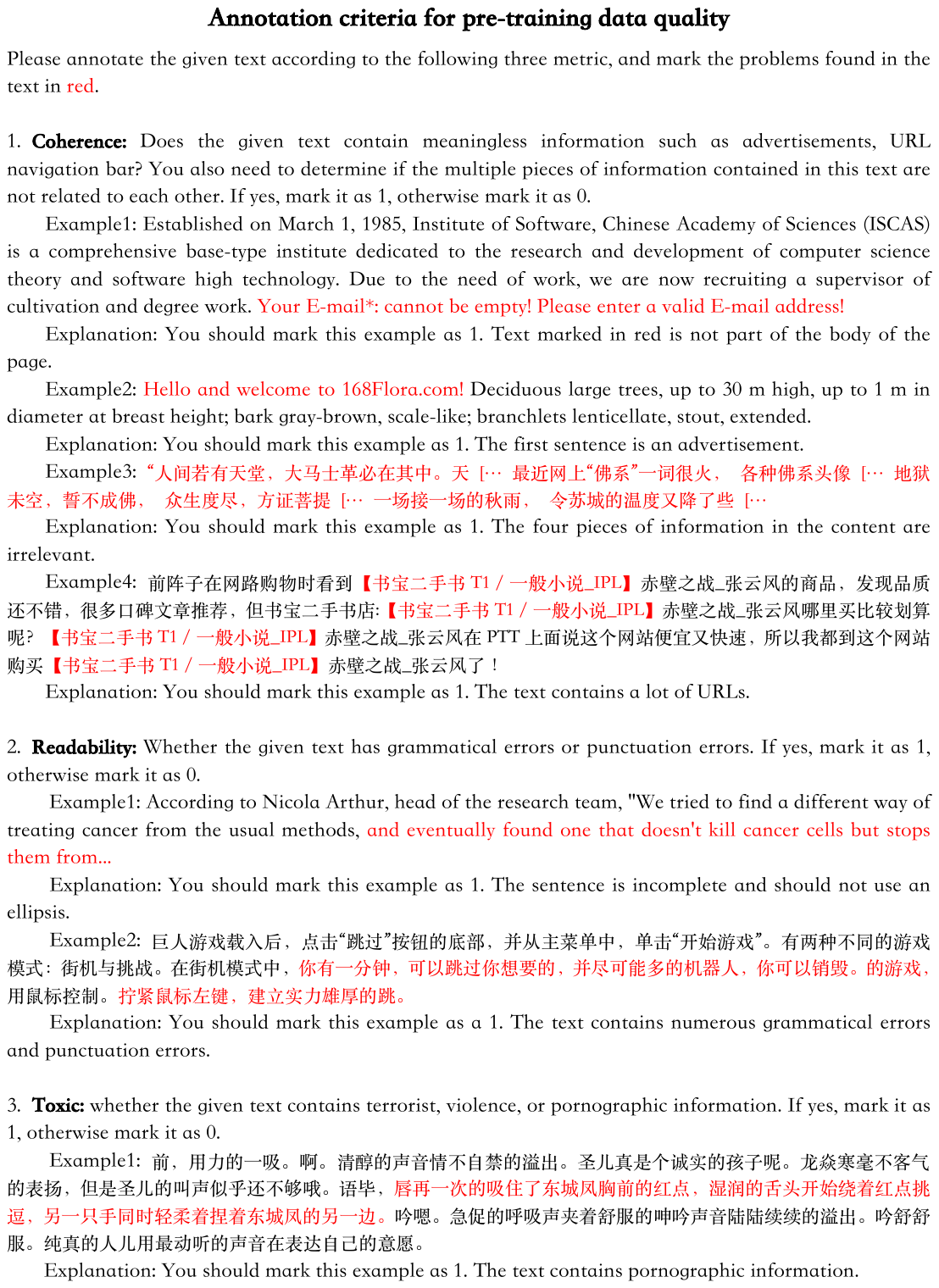}
    \caption{
    The criteria for human annotations.
    }
    \label{fig:annotations_criteria}
    \vspace{-0.3cm}
\end{figure*}

\clearpage
\onecolumn
\section{Pre-training Data}
\label{sec:pretraining_data}

The following charts presents examples of unsupervised and supervised training data.

\tcbset{width=1\textwidth,    
    colbacktitle = red!10!green!30!blue!80!,
    coltitle = white!90!white,
    fonttitle=\bfseries\large,
    colback=white!10!white,
    subtitle style={boxrule=0pt,colback=yellow!50!white}
    }
\lstset{language=sql, breaklines=true}

\begin{CJK*}{UTF8}{gbsn}

\begin{tcolorbox}[title=English Unsupervised Data]
\small
\centering
\setlength{\tabcolsep}{0.5mm}{
\begin{xltabular}{\linewidth}{
    X
    }
\normalsize{\color{black}\textbf{Text}} \\\\
\emph{ABC News’ Good Morning America outstripped NBC News’ Today by 761,000 viewers and 279,000 news demo viewers the week of April 7. It’s GMA‘s seventh consecutive week on top of the morning infotainment show race in both metrics, and its largest demo margin in three months. GMA has ranked No. 1 in overall audience for 89 of the past 93 weeks, and No. 1 in the news demo for 25 of this season’s 29 weeks to date. Today meanwhile, boasted it finished first with the younger, 18-49 year old age bracket, for the 42nd consecutive week. Today is on top of the ratings in the day part with men 25-54 this season, NBC noted — as well as adults, men and women 18-49. Today has posted seven consecutive months of ratings growth in total viewers, and both the 25-54 and 18-49 demos which NBC says is the show’s biggest ratings uptick since ’97. For the week, GMA clocked 5.617 million viewers — 2.212 million in the demo. Today logged 4.856 million viewers — 1.933 million in the demo. GMA bested CBS This Morning‘s 3.041 million viewers — 956,000 in the news demo.}
\end{xltabular}
}
\end{tcolorbox}

\vspace{0.1em}
\begin{tcolorbox}[title=Chinese Unsupervised Data]
\small
\centering
\setlength{\tabcolsep}{0.5mm}{
\begin{xltabular}{\linewidth}{
    X
    }
\normalsize{\color{black}\textbf{Text}} \\\\
\textit{你走了，带着我们还来不及开始的爱。你说我不够勇敢与果断、说我只是把喜欢寄放在你那边却从来没有想细心灌溉。但我要怎么能灌溉呢？你就像是一沃富饶的土地，大家总把他们的爱埋在你的心窝。圳水引来的滋润来不及洒落在我需要你的那些岁月，我向着有你在的阳光趋去，却如伊藤润二笔下的漩涡，只不过成了让人害怕的扭曲怪状。\textbf{\{...\}} 所以谢谢我曾经在你心里的位置那么前面、那么靠近你心的地方。}\\ \\
\textit{You're gone, with the love we haven't had time to start yet. You say I am not brave and decisive enough, and you say I just place my love on your side but never want to irrigate it carefully. But how can I irrigate it? You are like a fertile land, where everyone always buries their love in your heart. The nourishment brought by the water in the canal did not fall on the years when I needed you. I moved towards the sunshine with you, but like a vortex in the pen of Runji Ito, it only became a frightening twisted and strange shape \textbf{\{...\}} So thank you for being so in front of you and so close to your heart.}
\end{xltabular}
}
\end{tcolorbox}
\vspace{0.1em}

\begin{tcolorbox}[title=English Supervised Data]
\small
\centering
\setlength{\tabcolsep}{0.5mm}{
\begin{xltabular}{\linewidth}{
    X
    }
\normalsize{\color{black}\textbf{Text}} \\\\
\textit{Problem:}\\\textit{Human rights are moral principles or norms, which describe certain standards of human behaviour\textbf{\{...\}} Military aircraft can be either combat or non-combat: TurboJET is the brand name for the operations of the Hong Kong-headquartered Shun Tak-China Travel Ship Management Limited, which was established from the joint venture between Shun Tak Holdings Limited and China Travel International Investment Hong Kong Limited in July 1999. It operates hydrofoil ferry services in southern China. Question:use beechcraft starship} \\  \textit{A.military}\\ \textit{B.general aviation}\\ \textit{C.service}\\ \textit{Answer:B.general aviation }
\end{xltabular}
}
\end{tcolorbox}
\vspace{0.1em}

\begin{tcolorbox}[title=Chinese Supervised Data]
\small
\centering
\setlength{\tabcolsep}{0.5mm}{
\begin{xltabular}{\linewidth}{
    X
    }
\normalsize{\color{black}\textbf{Prompt}} \\ \\ 
问题:\{问题和内容\} 回答:\{回答的内容\}\\ \\
Question:\{the question\} Answer:\{the answer\}\\ \\
\normalsize{\color{black}\textbf{Text}} \\ \\
\textit{问题：}\\
\textit{请问下面描述属于哪一种事件类型？ 文章：昨晚，在号称``亚洲第一魔鬼主场''的天河体育场，国足在占据天时地利人和的情况下，半场就击垮对手，狂轰6球 \textbf{\{...\}} 这样的战绩和表现也在赛后引发了巨大的争议。}\\
\textit{A.胜负} \\\textit{B.加息}\\ \textit{C.融资}\\ \textit{D.爆炸} \\
\textit{回答：答案：A.胜负}\\ \\
\textit{Question:}\\
\textit{May I ask which type of event does the following description belong to? Article: Last night, at the Tianhe Stadium, which is known as the "Number One Devil's Stadium in Asia," the Chinese football team occupied Under the favorable conditions of timing, location, and people, he defeated his opponent in half time and roared 6 goals \textbf{\{...\}} Such achievements and performances have also sparked great controversy after the game.}\\
\textit{A.Success or Failure} \\\textit{B.Raise Interest Rates}\\ \textit{C.Financing}\\ \textit{D.Explosion} \\  
\textit{Answer: The answer is：A.Success or Failure}
\end{xltabular}
}
\end{tcolorbox}
\vspace{0.1em}

\begin{tcolorbox}[title=Code]
\small
\centering
\setlength{\tabcolsep}{0.5mm}{
\begin{xltabular}{\linewidth}{
    X
    }
\normalsize{\color{black}\textbf{Prompt}} \\\\ 

\texttt{\`}\texttt{\`}\texttt{\`}  (sql / python /  c / etc.)  $+$ \{Code\}  \texttt{\`}\texttt{\`}\texttt{\`} \\\\

\normalsize{\color{black}\textbf{Text}} \\\\

\textit{\texttt{\`}\texttt{\`}\texttt{\`} SQL }\\
\textit{\lstinline{ALTER TABLE kooste.tamperewfs_luontorastit DROP CONSTRAINT tamperewfs_luontorastit_pk;}} \\
\textit{\lstinline{ALTER TABLE kooste.tamperewfs_luontorastit DROP CONSTRAINT tunnus_rasti_unique;}}\\ 
\textit{\lstinline{ALTER TABLE kooste.tamperewfs_luontorastit ADD COLUMN kooste_id bigint NOT NULL GENERATED ALWAYS AS IDENTITY;}...\{Code\textasciitilde\} } \\
\textit{\texttt{\`}\texttt{\`}\texttt{\`}}
\end{xltabular}
}
\end{tcolorbox}
\vspace{0.1em}

\begin{tcolorbox}[title=MetaMath]
\small
\centering
\setlength{\tabcolsep}{0.5mm}{
\begin{xltabular}{\linewidth}{
    X
    }
\normalsize{\color{black}\textbf{Prompt}} \\\\

Question/Q:\{the question\} Answer/A:\{the answer\}\\\\

\normalsize{\color{black}\textbf{Text}} \\\\

\textit{Question: } \\
\textit{The coffee shop sold 8 more than four times the number of lattes as it did teas. If 6 teas were sold how many lattes did the coffee shop sell?}\\
\textit{Answer:} \\
\textit{If the coffee shop sold 6 teas, then the number of lattes sold is 8 more than four times the number of teas. Four times the number of teas is 4 * 6 = 24. So, the number of lattes sold is 24 + 8 = 32. The answer is: 32} \\ 
\end{xltabular}
}
\end{tcolorbox}
\vspace{0.1em}

\end{CJK*}

\section{Generation Cases}
\label{sec:generation_cases}
The following charts presents examples generated by Ziya2 with particular inputs.
\begin{CJK*}{UTF8}{gbsn}
\tcbset{width=1\textwidth,    
    colbacktitle = red!10!green!30!blue!80!,
    coltitle = white!90!white,
    fonttitle=\bfseries\large,
    colback=white!10!white,
    subtitle style={boxrule=0pt,colback=red!10!green!30!blue!40!}
    }
\begin{tcolorbox}[title=Chinese Generation]
\small
模型输入: \textit{我国的三皇五帝分别指的是谁？}\\\\
Model Input: \textit{Who are the ``Three Emperors and Five Sovereigns'' of our country?}
\tcbsubtitle{Ziya2 Generation}
\small
\centering
\setlength{\tabcolsep}{0.5mm}{
\begin{xltabular}{\linewidth}{
    X
    }
 "三皇五帝"是指我国古代部落首领，其中"三皇"一般认为指的是燧人、伏羲、神农；而"五帝"一般指的是黄帝、颛顼、帝喾、尧、舜。 \\\\
\textit{The "Three Emperors and Five Sovereigns" refer to ancient tribal leaders in China, among which the "Three Emperors" are generally believed to refer to Suiren, Fuxi, and Shennong; The "Five Sovereigns" generally refer to Huang Di, Zhuanxu, Sovereigns Ku, Yao, and Shun.} \\
\end{xltabular}
}
\end{tcolorbox}

\begin{tcolorbox}[title=English Generation]
\small
Model Input: \textit{Friends is a series of sitcoms produced by NBC in the United States,} 
\tcbsubtitle{Ziya2 Generation}
\small
\centering
\setlength{\tabcolsep}{0.5mm}{
\begin{xltabular}{\linewidth}{
    X
    }
\textit{ and created by David Crane and Marta Kauffman. The series premiered on September 22, 1994, and ended on May 6, 2004, lasting ten seasons. With an ensemble cast starring Jennifer Aniston, Courteney Cox, Lisa Kudrow, Matt LeBlanc, Matthew Perry and David Schwimmer, the show revolves around six friends in their 20s and 30s who live in Manhattan, New York City. The series was produced by Bright Kids Productions, in association with Warner Bros. Television. The original executive producers were Kevin S. Bright, Marta Kauffman, and David Crane.
Friends had received both critical acclaim and popular success throughout its original ten-season run on NBC; it received accolades such as winning the Emmy Award for Outstanding Comedy Series in 2002, and was ranked No. 25 on TV Guide's 50 Greatest TV Shows of All Time in 2002.} \\
\end{xltabular}
}
\end{tcolorbox}

\begin{tcolorbox}[title=Code Generation]
\small
模型输入: \textit{给定一个整数 n，计算所有小于等于 n 的非负整数中数字 1 出现的个数。例如：}
\\\textit{输入：n = 13 }
\\\textit{输出：6}
\\\textit{分析如下：输入n=13，小于等于13非负整数里面包含1的数字为1，10，11，12，13，所以数字1共出现了6次。下面是具体的python代码：
\texttt{\`}\texttt{\`}\texttt{\`}python}
\\
\\Model Input: \textit{Given an integer n, calculate the number of occurrences of the number 1 in all non negative integers less than or equal to n. For example:}
\\\textit{Input: n=13}
\\\textit{Output: 6}
\\\textit{The analysis is as follows: input n=13, and the number containing 1 in a non negative integer less than or equal to 13 is 1, 10, 11, 12, 13, so the number 1 appears together Six times. The following is the specific Python code: \texttt{\`}\texttt{\`}\texttt{\`} Python}
\lstset{language=python, breaklines=true}

\tcbsubtitle{Ziya2 Generation}
\small
\centering
\setlength{\tabcolsep}{0.5mm}{
\begin{xltabular}{\linewidth}{
    X
    }
\makecell[Xt]{\lstinline{import re} \\
\qquad\lstinline{def countDigitOne(self, n: int) -> int:}\\
\qquad\qquad\lstinline{count = 0}\\
\qquad\qquad\lstinline{for i in range(1, n + 1):}\\
\qquad\qquad\qquad\lstinline{s = str(i)}\\
\qquad\qquad\qquad\lstinline{for x in s:}\\
\qquad\qquad\qquad\qquad\lstinline{if x == '1':}\\
\qquad\qquad\qquad\qquad\qquad\lstinline{count += 1}\\
\qquad\lstinline{return count}\\
\texttt{\`}\texttt{\`}\texttt{\`}
}

\end{xltabular}
}
\end{tcolorbox}

\end{CJK*}

\end{document}